\documentclass[runningheads]{llncs}
\usepackage[T1]{fontenc}
\usepackage{graphicx}
\usepackage[inline]{enumitem}

\usepackage{tikz}
\usepackage[acronym]{glossaries}
\usepackage{glossaries-extra}
\usepackage[super]{nth}
\setabbreviationstyle[acronym]{long-short}
\newacronym{ai}{AI}{Artificial Intelligence}
\newacronym{ml}{ML}{Machine Learning}
\newacronym{rl}{RL}{Reinforcement Learning}
\newacronym{dl}{DL}{Deep Learning}
\newacronym{ot}{OT}{Operational Technology}
\newacronym{plc}{PLC}{Programmable Logic Controller}
\newacronym{hil}{HiL}{Hardware-in-the-Loop}
\newacronym{ioe}{IoE}{Internet of Everything}
\newacronym{ppo}{PPO}{Proximal Policy Optimization}
\newacronym{dqn}{DQN}{Deep Q-Learning}
\newacronym{opcua}{OPC~UA}{Open Platform Communication Unified Architecture}
\newacronym[plural=pn,firstplural=Petri Nets (PNs)]{pn}{PN}{Petri Net}
\newacronym[plural=ics,firstplural=Industrial Control Systems (ICS)]{ics}{ICS}{Industrial Control System}
\newacronym[plural=cps,firstplural=Cyber-Physical Systems (CPS)]{cps}{CPS}{Cyber-Physical System}
\usepackage{cleveref}
\crefname{section}{Sect.}{Sect.}
\Crefname{section}{Section}{Sections}
\crefname{figure}{Fig.}{Fig.}

\begin{document}
\title{A Modular Test Bed for Reinforcement Learning Incorporation into Industrial Applications\thanks{Reuf Kozlica and Simon Hirländer are supported by the Lab for Intelligent Data Analytics Salzburg (IDA Lab) funded by Land Salzburg (WISS 2025) under project number 20102-F1901166-KZP. Georg Schäfer is supported by the JRC ISIA project funded by the Christian Doppler Research Association.}}
%
%
\author{Reuf Kozlica\inst{1}\orcidID{0000-0003-1872-1685} \and
Georg Schäfer\inst{1}\orcidID{0009-0004-7622-7401} \and
Simon Hirländer\inst{2}\and
Stefan Wegenkittl\inst{1}\orcidID{0000-0002-3297-7997}}
\authorrunning{R. Kozlica et al.}
\titlerunning{A Modular Test Bed for RL Incorporation into Industrial Applications}
%
\institute{Salzburg University of Applied Sciences, Salzburg, Austria \\
\email{\{reuf.kozlica, georg.schaefer, stefan.wegenkittl\}@fh-salzburg.ac.at} \and
Paris Lodron University Salzburg, Salzburg, Austria \\
\email{simon.hirlaender@plus.ac.at}}
\maketitle              
\begin{abstract}
This application paper explores the potential of using reinforcement learning (RL) to address the demands of Industry~4.0, including shorter time-to-market, mass customization, and batch size one production.
Specifically, we present a use case in which the task is to transport and assemble goods through a model factory following predefined rules. Each simulation run involves placing a specific number of goods of random color at the entry point. The objective is to transport the goods to the assembly station, where two rivets are installed in each product, connecting the upper part to the lower part.
Following the installation of rivets, blue products must be transported to the exit, while green products are to be transported to storage. The study focuses on the application of reinforcement learning techniques to address this problem and improve the efficiency of the production process. 

\keywords{Reinforcement Learning \and Industry~4.0 \and OPC~UA.}
\end{abstract}
\section{Introduction} \label{sec:introduction}
Schäfer et al. stress how \gls{rl} can be used to overcome demands posed by the concepts of Industry~4.0, including shorter time-to-market, mass customization of products, and batch size one production proposing an \gls{ot}-aware \gls{rl} architecture in~\cite{schaefer2023deploying_rl}. \gls{rl} is an important machine learning paradigm for Industry~4.0 because it has the potential to surpass human level performance in various complex tasks~\cite{nian2020rlreview} and does not require pre-generated data in advance. \gls{rl} agents can learn optimal policies for executing control tasks, potentially leading to productivity maximization and cost reduction~\cite{kober2013reinforcement}. \gls{rl} agents can also explore their environment to generate new data, which is particularly useful in environments where data is scarce. Additionally, \gls{rl} agents can exploit their environment to detect unexpected behavior early on, supporting the creation of more realistic digital representations of the environment. For more information on \gls{rl} refer to Sutton \& Barto~\cite{sutton2018reinforcement}.

Based on a quantitative text analysis and a qualitative literature review, Hermann et al.~\cite{hermann2016industry} identified the following four Industry~4.0 design principles, which we are addressing with the created test bed: \begin{enumerate*}[label=(\roman*)]
  \item \label{enum:decentralized_decisions} Decentralized Decisions,
  \item \label{enum:technical_assistance} Technical Assistance,
  \item \label{enum:interconnection} Interconnection and
  \item \label{enum:information_transparency} Information Transparency.
\end{enumerate*}
\ref{enum:decentralized_decisions} and \ref{enum:technical_assistance} describe the interconnection of objects and people which allows for decentralized decision-making in Industry~4.0 enabled by \glspl{cps}. Meanwhile, humans' role is shifting towards strategic decision-making and problem-solving, supported by assistance systems and physical support by robots. Including \gls{rl} into the aforementioned setting allows for incorporation of both principles. \ref{enum:interconnection} and \ref{enum:information_transparency} address the increasing number of interconnected objects and people in the \gls{ioe} which enables collaborations and information transparency, but also requires common communication standards and cybersecurity, and relies on context-aware systems for appropriate decision-making based on real-time information provision. Combining \gls{opcua}~\cite{mahnke2009opc} with the standard \gls{rl} setting, as we are proposing in~\cite{schaefer2023deploying_rl}, enables the support for design principles~\ref{enum:interconnection} and~\ref{enum:information_transparency}.

\section{Test Bed}
\begin{figure*}[ht]
    \centerline{\includegraphics[width=0.8769420\linewidth]{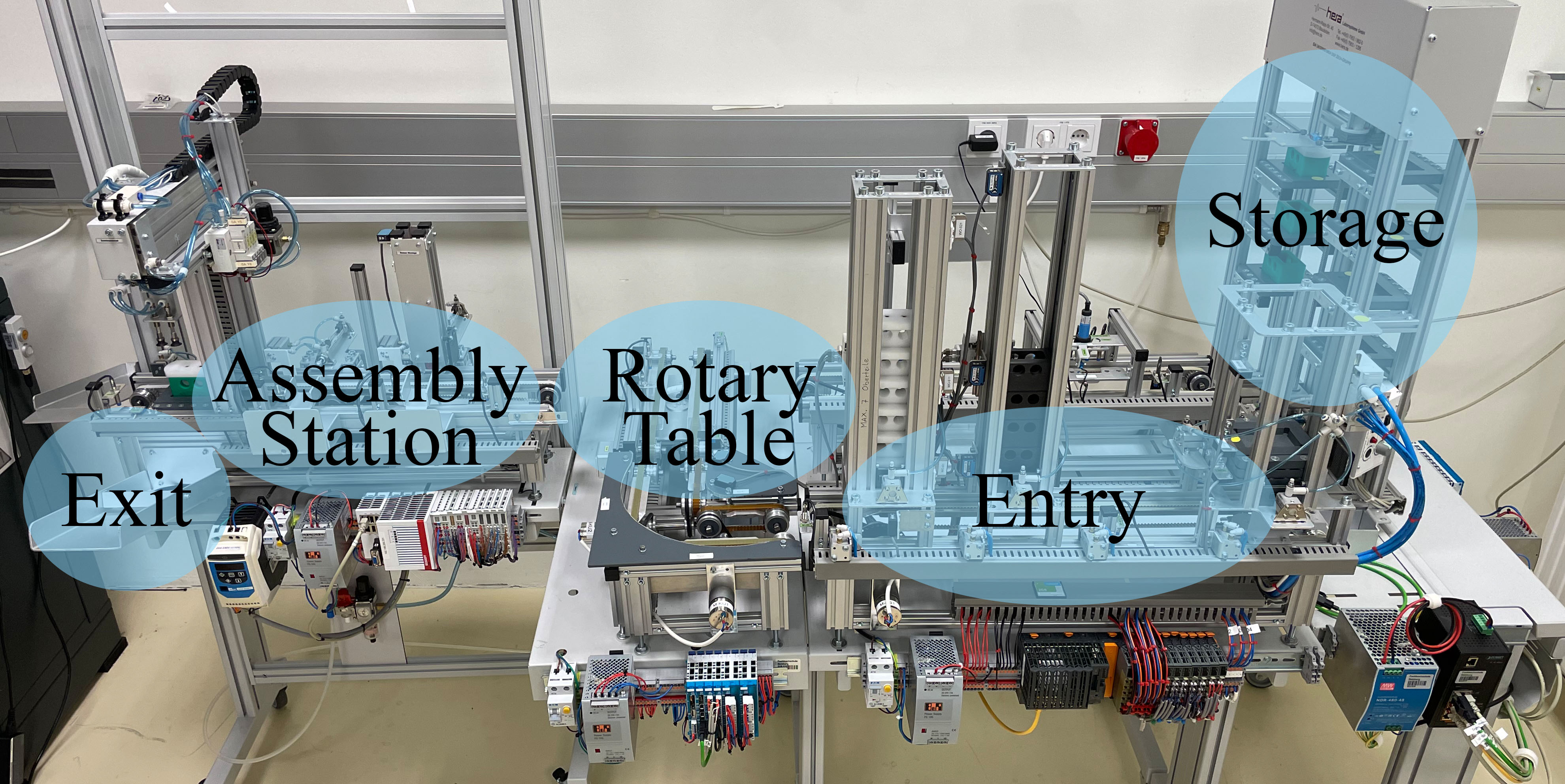}}
    \caption[Model factory placed in the Smart Factory Lab]{Model factory placed in the Smart Factory Lab\footnotemark.}
    \label{fig:model_factory}
\end{figure*}

The case study presented by \cite{riedmann2022petrinetsimulation} and extended by \cite{harb2022strategies} aims to simulate a real production system using a model factory. The model factory, depicted in \cref{fig:model_factory}, comprises five modules: entry, rotary table, assembly station, storage, and exit. The entry storage of the model factory stores three different types of parts, namely the transport carriage, lower part of the product, and upper part of the product. The transport carriage is used for moving goods on the conveyor belts, while the lower and upper parts of the product are required for product assembly. 
The rotary table, which serves as a pivotal element in the model factory, can transport goods from the entry to the storage unit or from the entry to the assembly station. At the assembly station, rivets must be added to the product, and during the insertion process, the conveyor must remain stationary. The assembled products come in different randomly assigned colors. The factory is also equipped with several sensors that track the products on their carriages as they move through the facility, and can recognize the color and presence of rivets in a product.

\footnotetext{\url{https://its.fh-salzburg.ac.at/forschung/smart-factory-lab/}}

The model is designed to be representative of a real production system and includes several important aspects such as transportation of goods using conveyor belts and a rotary table, product modification through the assembly station, and storage of goods in both entry point and the main storage unit. Moreover, it represents a modular production plant consisting of multiple \glspl{plc} of different manufacturers.

\subsection{OPC~UA based RL-OT Integration}
Addressing the principles \ref{enum:interconnection} and \ref{enum:information_transparency} introduced in \cref{sec:introduction}, \cite{schaefer2023deploying_rl} proposes an \gls{opcua} based architecture for \gls{rl} in the context of industrial control systems. \gls{opcua} allows communication between various industrial devices, including both real and simulated devices. The proposed architecture uses \gls{opcua} nodes to extend the standard \gls{rl} setting. The mapping of the \gls{rl} action and state space with the \gls{opcua} address space is performed by the \gls{rl} mapper. The agent's action is turned into an \gls{opcua} call using a function that sets the corresponding actuators using the client-server model. The environment is notified of each sensor change using the PubSub mechanism, mapping \gls{opcua} sensor nodes to \gls{rl} states. After each sensor change, a reward evaluation is triggered and a state space transition occurs. The paper also discusses how custom object types can be created for nodes accessible to the RL agent to automate the mapping between the RL action and state space with the \gls{opcua} address space. This architecture allows for a seamless integration of different out-of-the-box implementations of \gls{rl} agents into \gls{ot} systems.

\subsection{Integration of Different \gls{rl} Agents}
Kozlica et al. present a simulation environment for a production line in which an agent controlled by \gls{rl} algorithms transports and assembles goods \cite{kozlica2023dqnvsppo}. The goal is to test the performance of different \gls{rl} algorithms and their ability to solve the given task. 
Two different reward functions are defined and compared. The first reward function only focuses on correctly assembling and sorting the products, with no penalty for a simple transition. The second reward function also considers the number of transitions needed for task completion. Negative rewards are assigned to collisions, incorrectly sorted products, and invalid transitions. A positive reward is assigned for completing the task. Moreover, two different \gls{rl} algorithms are compared: \gls{dqn} and \gls{ppo}. The results show that both agents can learn to solve the production line task, with the \gls{ppo} agent generally outperforming the \gls{dqn} agent in terms of task completion and reward. The authors conclude that the presented simulation environment is suitable for testing and comparing different \gls{rl} algorithms for the production line task.

\section{Conclusion}
Our proposed test bed aims to demonstrate how \gls{rl} can be used to address the demands posed by Industry~4.0, particularly in relation to the four design principles decentralized decisions, technical assistance, interconnection, and information transparency. 
The test bed consists of a modular production plant and is used for different research based scenarios. In this paper, the authors have shown how to fulfill the Industry~4.0 design principles by incorporating an \gls{opcua} information model in the general \gls{rl} setting. Additionally, it was shown, that different already available \gls{rl} agent implementations can be used for solving the defined sorting task.

%
%
%
\bibliographystyle{splncs04}
\bibliography{references}

\end{document}